\documentclass[10pt, a4paper]{article}
\usepackage{listings}
\usepackage{xcolor}
\usepackage{float} 
\usepackage{amsmath}

\usepackage[final]{lrec2026} 
\title{LATA: A Tool for LLM-Assisted Translation Annotation}

\name{Baorong Huang, Ali Asiri\textsuperscript{*}\thanks{Corresponding author: \href{mailto:amaasiri@uqu.edu.sa}{amaasiri@uqu.edu.sa}}}   

\address{School of Foreign Languages, Huaihua University \\ Alith University College, Umm al-Qura University \\
         Huaihua, China, \\ Makkah, Saudi Arabia \\
         huangbaorong2021@gmail.com, amaasiri@uqu.edu.sa\\
         }

\abstract{
The construction of high-quality parallel corpora for translation research has increasingly evolved from simple sentence alignment to complex, multi-layered annotation tasks. This methodological shift presents significant challenges for structurally divergent language pairs, such as Arabic–English, where standard automated tools frequently fail to capture deep linguistic shifts or semantic nuances. This paper introduces a novel, LLM-assisted interactive tool designed to reduce the gap between scalable automation and the rigorous precision required for expert human judgment. Unlike traditional statistical aligners, our system employs a template-based Prompt Manager that leverages large language models (LLMs) for sentence segmentation and alignment under strict JSON output constraints. In this tool, automated preprocessing integrates into a human-in-the-loop workflow, allowing researchers to refine alignments and apply custom translation technique annotations through a stand-off architecture. By leveraging LLM-assisted processing, the tool balances annotation efficiency with the linguistic precision required to analyze complex translation phenomena in specialized domains.
 \\ \newline \Keywords{LLM-assisted alignment, parallel corpora, translation annotation} }

\begin{document}
\maketitleabstract
\section{Introduction}
The construction of parallel corpora has undergone a significant evolution, moving beyond the purely technical task of sentence alignment to a more sophisticated process involving multi-layered annotation of semantics and translation strategies\cite{abdullah2024comprehensive}. This development is driven by increasing demands in translation research and natural language processing (NLP) for corpora that do more than identify equivalent segments; they are expected to represent the translation process itself, including shifts in meaning, structural adjustments, and stylistic adaptations. This requirement is particularly pronounced in Arabic–English corpora, where substantial structural divergences, rich morphological systems, and domain-specific considerations render surface-level alignment inadequate for capturing deeper linguistic and translational phenomena.

To manage this growing complexity, stand-off annotation formats are increasingly adopted, such as BioNLP \cite{nedellec2013overview}, BRAT \cite{stenetorp2012brat}, and OneIE \cite{lin2020multilingual}. By separating annotation layers from the original text, these formats enable iterative refinement, the representation of nested and overlapping entities, and the application of multiple tags without altering the source material. This separation is important for expert validation, where maintaining data integrity is a fundamental requirement.

However, despite these improvements, there is still a gap in balancing the scalability of automation with the interpretive precision required for qualitative research. Rule-based automation struggles with the nuances of creative translation, while manual annotation is expensive for large-scale datasets. This paper introduces an LLM-assisted interactive tool designed to address this challenge. By combining template-based prompt engineering for segmentation and alignment with a human-in-the-loop post-editing workflow, our tool reduces the gap between the efficiency of generative AI and the fine-grained control required for rigorous academic annotation.

\begin{figure*}[!t]

  \centering

  \includegraphics[width=\textwidth]{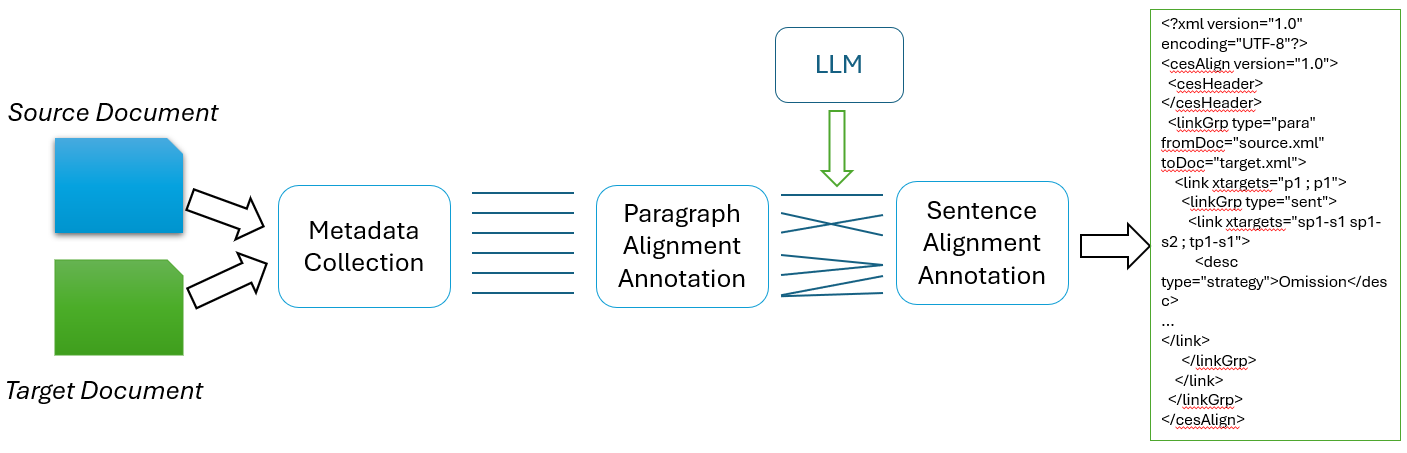}

  \caption{Hierarchical Alignment Pipeline. The process begins with metadata collection from source and target documents, followed by paragraph-level alignment. An LLM layer provides automated sentence segmentation and alignment to support the transition to granular sentence-level annotation. The final output is zipped three structured, CES-compliant XML files that contain alignment links and qualitative descriptions of translation techniques (see Appendix)}

  \label{fig:pipeline}

\end{figure*}

\subsection{Literature Review}

Current research identifies three distinct paradigms for corpus annotation, each offering different trade-offs between computational scalability and linguistic precision. Manual annotation remains the gold standard for quality, particularly for structurally divergent language pairs. While tools like the LDC aligner\cite{grimes2010creating} and PROJECTOR\cite{akbik2017projector}  provide intuitive annotation GUI and guidelines, the process is labor-intensive. In contrast, automatic pre-annotation can accelerate workflows, but the cognitive load required to inspect and correct error-prone outputs often offsets these gains.

Conversely, fully automatic approaches prioritize scalability by "projecting" annotations from resource-rich to under-resourced languages via word alignment. Frameworks such as ZAP\cite{akbik2018zap} and MultiSemCor \cite{bentivogli2005exploiting} demonstrate high efficiency for low-level tasks like POS tagging and NER, achieving precision up to 86\% \cite{bentivogli2004evaluating}. However, these methods falter as the depth of the annotation increases; non-literal translations and phrasal mismatches frequently result in errors that require substantial manual correction, limiting their use for complex semantic analysis \cite{ehrmann2011building, hammouda2016integration}.

A third, intermediate approach consists of interactive visualization and human-in-the-loop tools. Systems such as BRAT\cite{stenetorp2012brat} , AlvisAE \cite{papazian2012alvisae}, TextAE\cite{grundke2016textai}, Fassieh \cite{attia2009fassieh}, and InterText\cite{vondvrivcka2016intertext} exemplify environments where automation supports analysis and pre-processing, while humans retain control over validation and correction. These tools emphasize segment-level operations, visual diagnostics, and stand-off architectures, allowing annotators to manage annotation errors, ambiguity, and structural mismatches effectively.

Building on these approaches, our approach combines LLM-driven segmentation and alignment with a visual interface for manual control, preserving automation’s efficiency while ensuring the linguistic rigor and control essential for scholarly annotation.

\section{Proposed Translation Annotation Pipeline}
The proposed approach consists of a three-phase translation annotation pipeline. It turns the raw textual data and high-quality, machine-readable translation corpora by leveraging a hierarchical refinement process. The three phases are Document Metadata Collection, Paragraph Alignment Annotation, and LLM-Assisted Sentence Segmentation and Alignment Annotation (see Figure \ref{fig:pipeline}).
\subsection{Metadata Collection}
The first phase involves the extraction and tagging of "extralinguistic" data, including document title, author, publication, publication time, publisher, domain, language, and other information. Prior to the alignment, this phase records the document’s distinctive features as document metadata. Let $D$ represent the document. Metadata $M$ includes $M = \{author, genre, date, source\_language, \\ target\_language\}$. This phase provides the "labels" that support targeted, filtered research. For instance, allowing researchers to isolate texts by a specific author, compare works within a particular genre, and analyze translations from a certain time period.
\subsection{Paragraph Alignment Annotation}
Once the document metadata is collected, the pipeline moves on to the Paragraph Alignment Annotation and supports to align paragraphs, the primary structural units.
Let the source document be a set of paragraphs $P_{src} = \{p_1, p_2, ..., p_n\}$ and the target document be $P_{tgt} = \{p'_1, p'_2, ..., p'_m\}$.
This phase establishes a mapping $f: p_i \to p'_j$. Each pair $(p_i, p'_j)$ may be accompanied by a comment ($C$) to form $A_p = (p_i, p'_j, C)$ The comment $C$ allows annotators to note structural deviations, such as when one source paragraph is split into two in the target language for stylistic flow ($p_1 \to \{p'_1, p'_2\}$).
\subsection{LLM-Assisted Sentence Segmentation, Alignment and Annotation}
The final phase is the most granular, utilizing LLMs to handle the complex task of sentence-level mapping within the already-aligned paragraphs.
For a given paragraph $p_i$, this phase identifies sentences $s$:$$p_i = \{s_{i,1}, s_{i,2}, ..., s_{i,k}\}$$ An LLM is prompted to suggest the most likely alignment between $s_{i,k}$ (source sentences) and $s'_{j,l}$ (target sentences). It supports 1:1, 1:N, M:N, or N:1 mappings. The human annotator reviews the output of the LLM. If the source sentence $p_1\text{-}s_1$ corresponds to the target sentence $p'_1\text{-}s_1$, the alignment is saved into the embedded database for persistence.

\section{Description of LATA: A Desktop Application}
\subsection{Architecture}
The technical framework of the tool is built upon a decoupled React-Electron-SQLite stack, engineered to balance high-performance text processing with future extensibility. The frontend is developed using React, providing a modular and responsive environment tailored to the demands of translation annotation. The Electron main process serves as the system’s core, which uses  Inter-Process Communication (IPC) to bridge the high-level user interface with underlying data services and local system resources. Data storage is maintained via a SQLite database, which offers a lightweight yet powerful local storage solution.

\subsection{Features}
\subsubsection{Customizing Prompt}
\begin{figure}[H]
  \centering
  \includegraphics[width=\columnwidth]{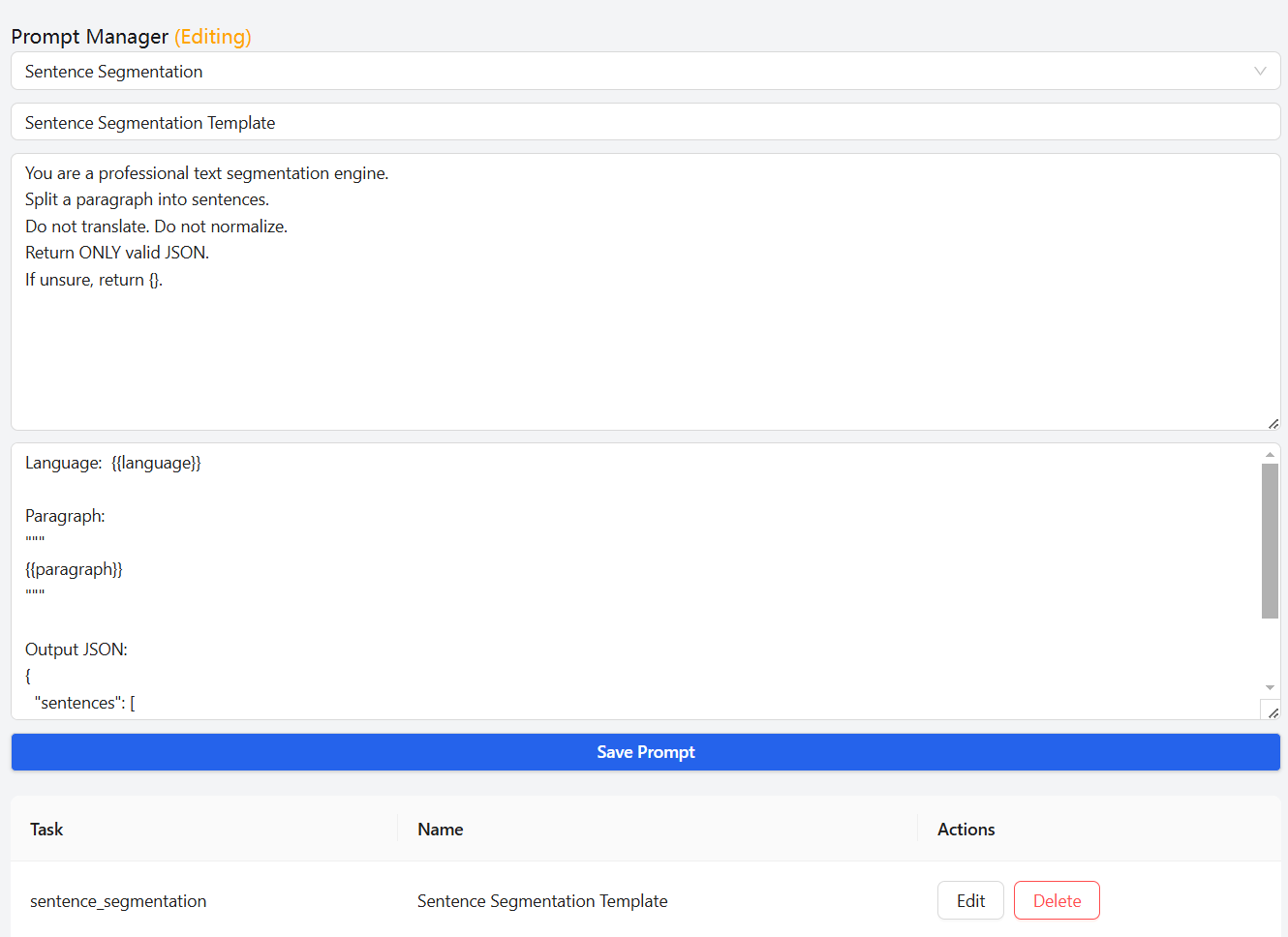}
  \caption{Template-based prompt configuration with dynamic placeholders}
  \label{fig:promptmanager}
\end{figure}

The tool utilizes a template-based prompt system that enables users to use the placeholders, such as {{language}} and {{paragraph}}, within the instruction template, as shown in Figure \ref{fig:promptmanager}. By decoupling core alignment logic from variable input data, this approach provides more flexibility. Furthermore, the templates enforce a strict JSON output schema, guaranteeing that segmented text is consistently formatted for the downstream processing pipelines.
\subsubsection{Customizing Technique Annotation}

\begin{figure}[H]
  \centering
  \includegraphics[width=\columnwidth]{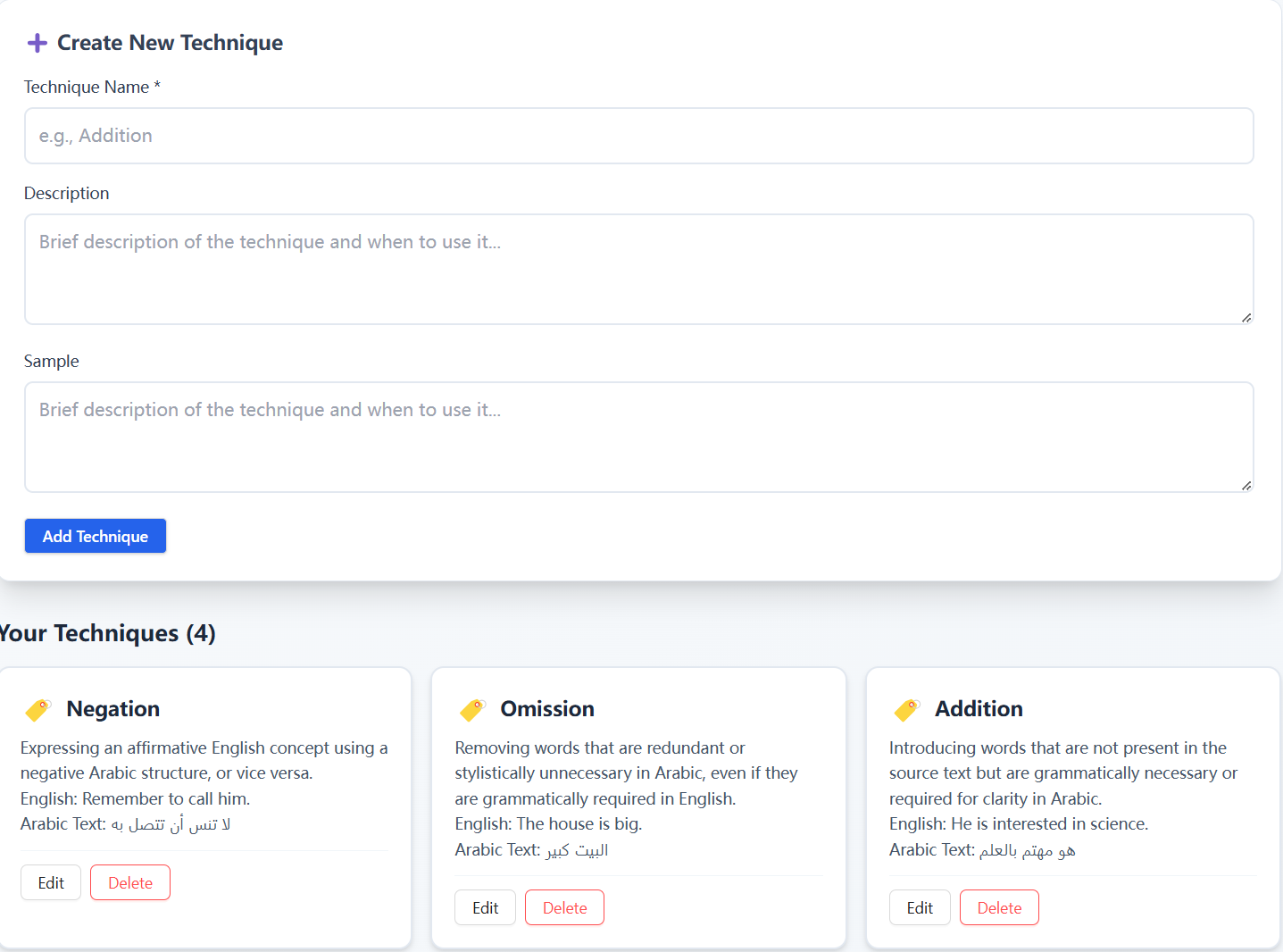}
  \caption{Translation annotation configuration with customized name, description, and examples}
  \label{fig:techniquemanager}
\end{figure}

The tool supports the customization of translation technique annotations, allowing researchers to establish a customized taxonomy for corpus analysis. Users can define specific categories by entering a technique name, such as Negation, Omission, and Addition,  a theoretical description, and illustrative  examples, as shown in Figure \ref{fig:techniquemanager}. This feature ensures that the annotation schema aligns with the specific research framework, facilitating the systematic identification of translation shifts within the parallel text.

\subsubsection{Defining Metadata}
This feature allows researchers to specify bibliographic details of both source and target texts. Key attributes, such as domain, document type, and publisher, facilitate corpus categorization. This structured metadata collection ensures data provenance and enables variable-based sub-corpus analysis.
\begin{figure}[H]
  \centering
  \includegraphics[width=\columnwidth]{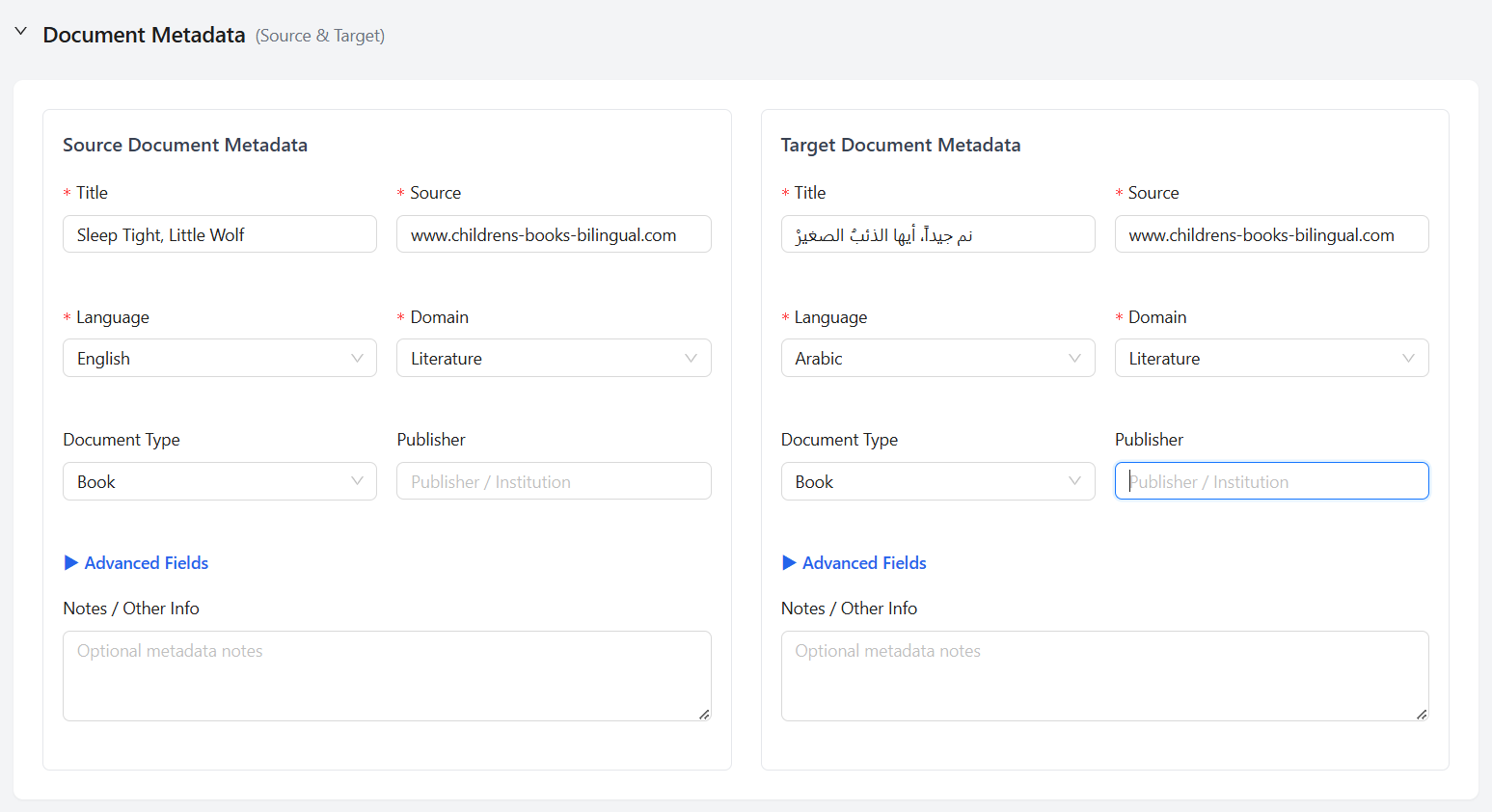}
  \caption{Dual-pane interface for capturing source and target document metadata}
  \label{fig:metadata}
\end{figure}
\subsubsection{Aligning Documents}
This feature serves as the post-editing environment for verifying and refining the initial alignments generated by the LLM. The dual-column layout displays source and target segments side-by-side, joined by interactive visual connectors that represent the current mapping status. Researchers can manually adjust these links to correct misalignment errors (handling 1:1, 1-to-many, many-to-many, or null matches) to ensure alignment quality. Additionally, the tool integrates the common editing operations, such as undo/redo, font configuration, and the modification of metadata. 
\begin{figure}[H]
  \centering
  \includegraphics[width=\columnwidth]{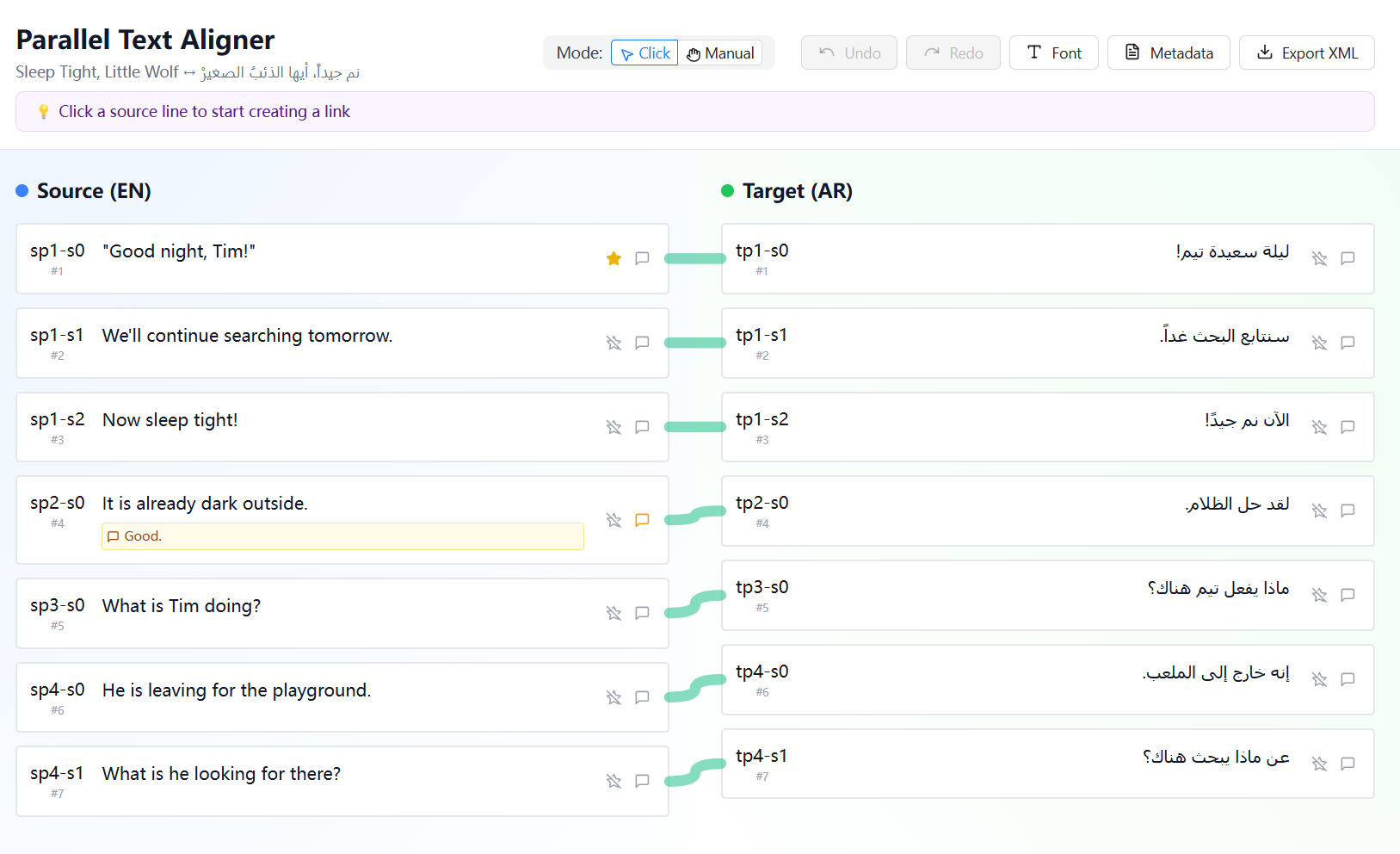}
  \caption{Manual alignment adjustment interface for post-editing parallel or sentence alignments}
  \label{fig:alignmentgui}
\end{figure}
\subsubsection{Annotating Translation Techniques}
The "Link Details" modal supports the granular annotation of individual alignment pairs. Activated by selecting a specific link, this interface allows researchers to categorize translation shifts using the pre-defined technique techniques (e.g., Inversion). A dedicated comment field supports qualitative analysis, enabling the recording of specific linguistic observations or rationales directly attached to each aligned segment pair.
\begin{figure}[H]
  \centering
  \includegraphics[width=\columnwidth]{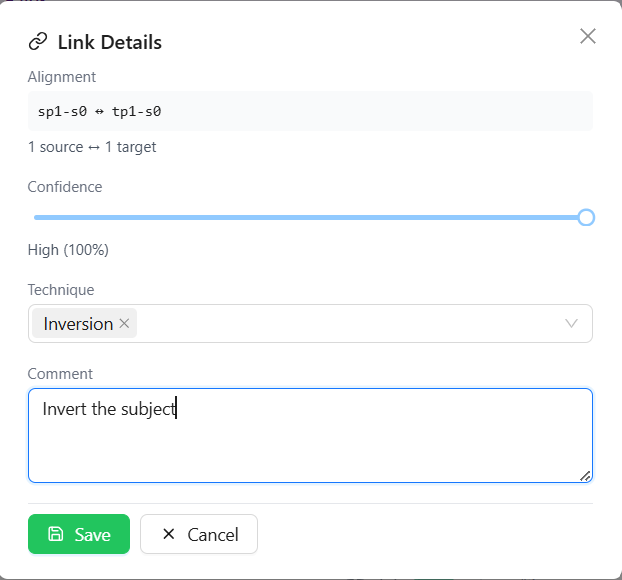}
  \caption{Translation technique annotation with structured categorization and qualitative comment support}
  \label{fig:techniqueannotation}
\end{figure}

\section{Aims of Further Development}
Further development of this pipeline extends its function beyond alignment into three interconnected layers: word-level annotation, knowledge graph construction, and multimodal anchoring.

First, at the micro level, the system enhances word-level alignment and annotation within Phase 3 sentence-level processing. Beyond segment alignment, it captures fine-grained linguistic shifts, including tense transformations, part-of-speech changes, lexical substitutions, and other morphosyntactic variations.

Second, the integration of NER facilitates the concurrent construction of a bilingual knowledge graph. Linguistic segments are mapped to structured ontological attributes, transforming aligned corpora into machine-readable knowledge representations. In this way, translation becomes a dual-purpose activity: it produces a validated target text while simultaneously populating a structured database of cultural and conceptual entities. This approach ensures that implicit cultural knowledge is preserved in an operable computational format.

Third, to address the increasing prominence of multimodal data, the tool will incorporate a visual–textual anchoring mechanism. This architecture enables researchers to link linguistic units at the paragraph level or sentence level to spatial coordinates within static images. Such functionality supports multimodal corpus analysis by documenting how visual semiotics influence translation strategies. For instance, when analyzing material related to intangible cultural heritage, annotators can anchor translated descriptions directly to the visual techniques depicted in an image. 

\section{Availability and License}
LATA is available under the MIT  License from
 \href{https://github.com/BaorongHuang-AI/lata.git}{Github}.



\section{Bibliographical References} 
\bibliographystyle{lrec2026-natbib}
\bibliography{mybib}

@article{abdullah2024comprehensive,
  title={A comprehensive review of existing corpora and methods for creating annotated corpora for event extraction tasks},
  author={Abdullah, Mohd Hafizul Afifi and Aziz, Norshakirah and Abdulkadir, Said Jadid and Hussain, Kashif and Alhussian, Hitham and Talpur, Noureen},
  journal={Journal of Data and Information Science},
  volume={9},
  number={4},
  pages={196--238},
  year={2024},
  publisher={De Gruyter Open Ltd.}
}

@inproceedings{akbik2017projector,
  title={The projector: An interactive annotation projection visualization tool},
  author={Akbik, Alan and Vollgraf, Roland},
  booktitle={Proceedings of the 2017 Conference on Empirical Methods in Natural Language Processing: System Demonstrations},
  pages={43--48},
  year={2017}
}

@inproceedings{nedellec2013overview,
  title={Overview of BioNLP shared task 2013},
  author={N{\'e}dellec, Claire and Bossy, Robert and Kim, Jin-Dong and Kim, Jung-Jae and Ohta, Tomoko and Pyysalo, Sampo and Zweigenbaum, Pierre},
  booktitle={Proceedings of the BioNLP shared task 2013 workshop},
  pages={1--7},
  year={2013}
}

@inproceedings{stenetorp2012brat,
  title={BRAT: a web-based tool for NLP-assisted text annotation},
  author={Stenetorp, Pontus and Pyysalo, Sampo and Topi{\'c}, Goran and Ohta, Tomoko and Ananiadou, Sophia and Tsujii, Jun’ichi},
  booktitle={Proceedings of the Demonstrations at the 13th Conference of the European Chapter of the Association for Computational Linguistics},
  pages={102--107},
  year={2012}
}

@inproceedings{akbik2018zap,
  title={ZAP: an open-source multilingual annotation projection framework},
  author={Akbik, Alan and Vollgraf, Roland},
  booktitle={Proceedings of the Eleventh International Conference on Language Resources and Evaluation (LREC 2018)},
  year={2018}
}

@inproceedings{ehrmann2011building,
  title={Building a multilingual named entity-annotated corpus using annotation projection},
  author={Ehrmann, Maud and Turchi, Marco and Steinberger, Ralf},
  booktitle={Proceedings of the International Conference Recent Advances in Natural Language Processing 2011},
  pages={118--124},
  year={2011}
}

@article{bentivogli2005exploiting,
  title={Exploiting parallel texts in the creation of multilingual semantically annotated resources: the MultiSemCor Corpus},
  author={Bentivogli, Luisa and Pianta, Emanuele},
  journal={Natural Language Engineering},
  volume={11},
  number={3},
  pages={247--261},
  year={2005},
  publisher={Cambridge University Press}
}

@inproceedings{bentivogli2004evaluating,
  title={Evaluating cross-language annotation transfer in the multisemcor corpus},
  author={Bentivogli, Luisa and Forner, Pamela and Pianta, Emanuele},
  booktitle={COLING 2004: Proceedings of the 20th International Conference on Computational Linguistics},
  pages={364--370},
  year={2004}
}

@inproceedings{hammouda2016integration,
  title={Integration of a segmentation tool for Arabic corpora in NooJ platform to build an automatic annotation tool},
  author={Hammouda, Nadia Ghezaiel and Haddar, Kais},
  booktitle={International NooJ Conference},
  pages={89--100},
  year={2016},
  organization={Springer}
}

@article{attia2009fassieh,
  title={Fassieh, a semi-automatic visual interactive tool for morphological, PoS-Tags, phonetic, and semantic annotation of Arabic text corpora},
  author={Attia, Mohamed and Rashwan, Mohsen AA and Al-Badrashiny, Mohamed ASAA},
  journal={IEEE transactions on audio, speech, and language processing},
  volume={17},
  number={5},
  pages={916--925},
  year={2009},
  publisher={IEEE}
}

@misc{vondvrivcka2016intertext,
  title={InterText editor v1. 5 comprehensive guide. Institute of the Czech National Corpus Charles University, Faculty of Arts},
  author={Vond{\v{r}}i{\v{c}}ka, P},
  year={2016}
}

@inproceedings{papazian2012alvisae,
  title={AlvisAE: a collaborative Web text annotation editor for knowledge acquisition},
  author={Papazian, Fr{\'e}d{\'e}ric and Bossy, Robert and N{\'e}dellec, Claire},
  booktitle={Proceedings of the Sixth Linguistic Annotation Workshop},
  pages={149--152},
  year={2012}
}

@inproceedings{grundke2016textai,
  title={TextAI: Enhancing TextAE with Intelligent Annotation Support.},
  author={Grundke, Maximilian and Jasper, Johannes and Perchyk, Mariya and Sachse, Jan Philipp and Krestel, Ralf and Neves, Mariana L},
  booktitle={SMBM},
  pages={80--84},
  year={2016}
}

@inproceedings{grimes2010creating,
  title={Creating arabic-english parallel word-aligned treebank corpora at LDC},
  author={Grimes, Stephen and Li, Xuansong and Bies, Ann and Kulick, Seth and Ma, Xiaoyi and Strassel, Stephanie},
  booktitle={Proceedings of Language Resources and Evaluation Conference (LREC’10), Malta},
  year={2010}
}

@inproceedings{lin2020multilingual,
  title     = {Multilingual Multitask Joint Neural Information Extraction},
  author    = {Lin, Ying and Ji, Heng and Li, Fei and Han, Xianpei and Huang, Shumin and Luo, Jie and Xu, Jingwei},
  booktitle = {Proceedings of the 58th Annual Meeting of the Association for Computational Linguistics (ACL)},
  year      = {2020},
  pages     = {1029--1040},
  publisher = {Association for Computational Linguistics},
  address   = {Online},
  doi       = {10.18653/v1/2020.acl-main.94},
  url       = {https://aclanthology.org/2020.acl-main.94}
}



\section{Language Resource References}
N/A
\bibliographystylelanguageresource{lrec2026-natbib}
\bibliographylanguageresource{languageresource}

\clearpage          
\onecolumn          
\section{Appendix}
\subsection{Sample Output in CES-compliant XML files}
\subsubsection{Alignment XML}
\begin{figure}[H]
  \centering
  \includegraphics[width=\columnwidth]{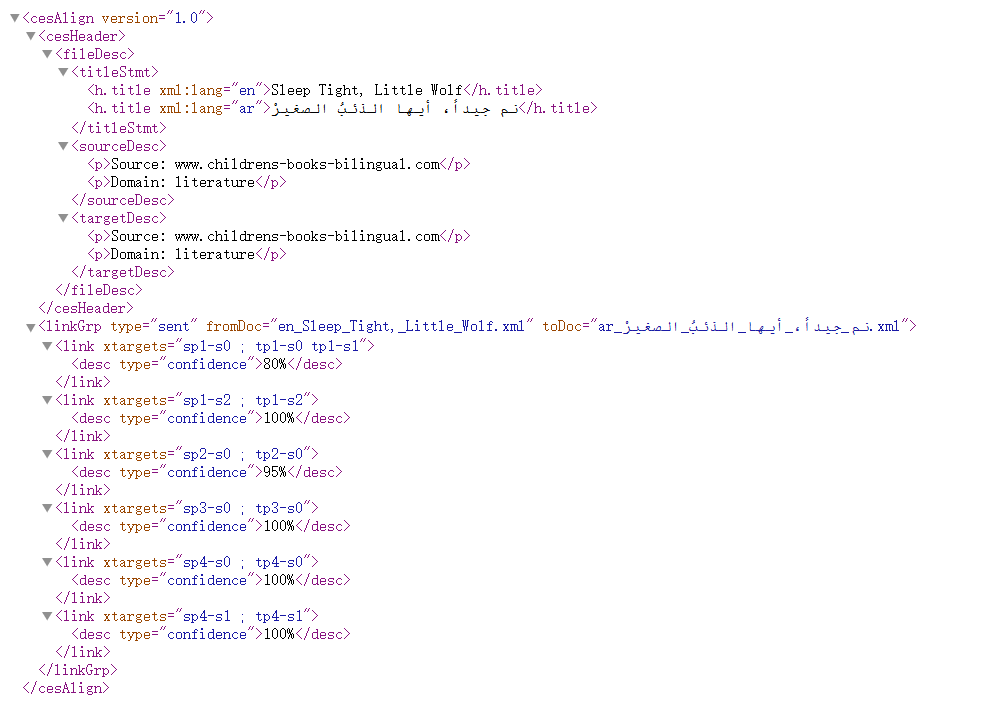}
  \caption{Sample alignment XML file}
  \label{fig:alignmentxml}
\end{figure}

\clearpage          
\onecolumn          
\subsubsection{Source XML}
\begin{figure}[H]
  \centering
  \includegraphics[width=\columnwidth]{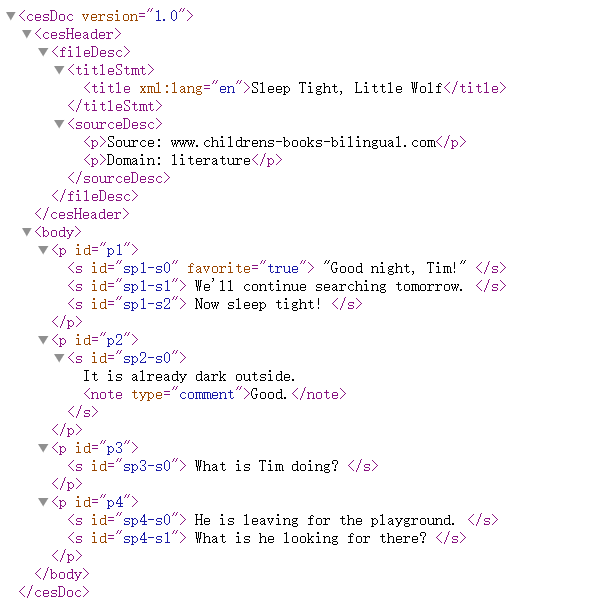}
  \caption{Sample source XML file}
  \label{fig:sourcexml}
\end{figure}


\clearpage          
\onecolumn          
\subsubsection{Target XML}
\begin{figure}[H]
  \centering
  \includegraphics[width=\columnwidth]{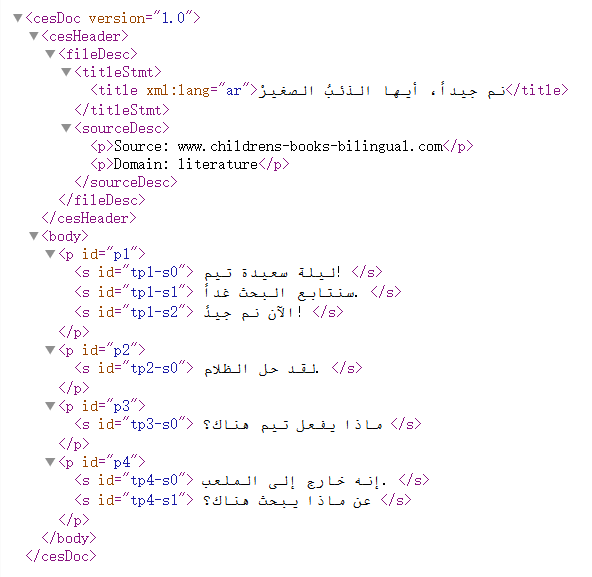}
  \caption{Sample target XML file}
  \label{fig:targetxml}
\end{figure}


\end{document}